# Fully Convolutional Neural Networks for Dynamic Object Detection in Grid Maps

Florian Piewak[1,2], Timo Rehfeld[1], Michael Weber[2,3] and J. Marius Zöllner[2,3]

***Abstract*— Grid maps are widely used in robotics to represent obstacles in the environment and differentiating dynamic objects from static infrastructure is essential for many practical applications. In this work, we present a methods that uses a deep convolutional neural network (CNN) to infer whether grid cells are covering a moving object or not. Compared to tracking approaches, that use e.g. a particle filter to estimate grid cell velocities and then make a decision for individual grid cells based on this estimate, our approach uses the entire grid map as input image for a CNN that inspects a larger area around each cell and thus takes the structural appearance in the grid map into account to make a decision. Compared to our reference method, our concept yields a performance increase from 83.9% to 97.2%. A runtime optimized version of our approach yields similar improvements with an execution time of just 10 milliseconds.**

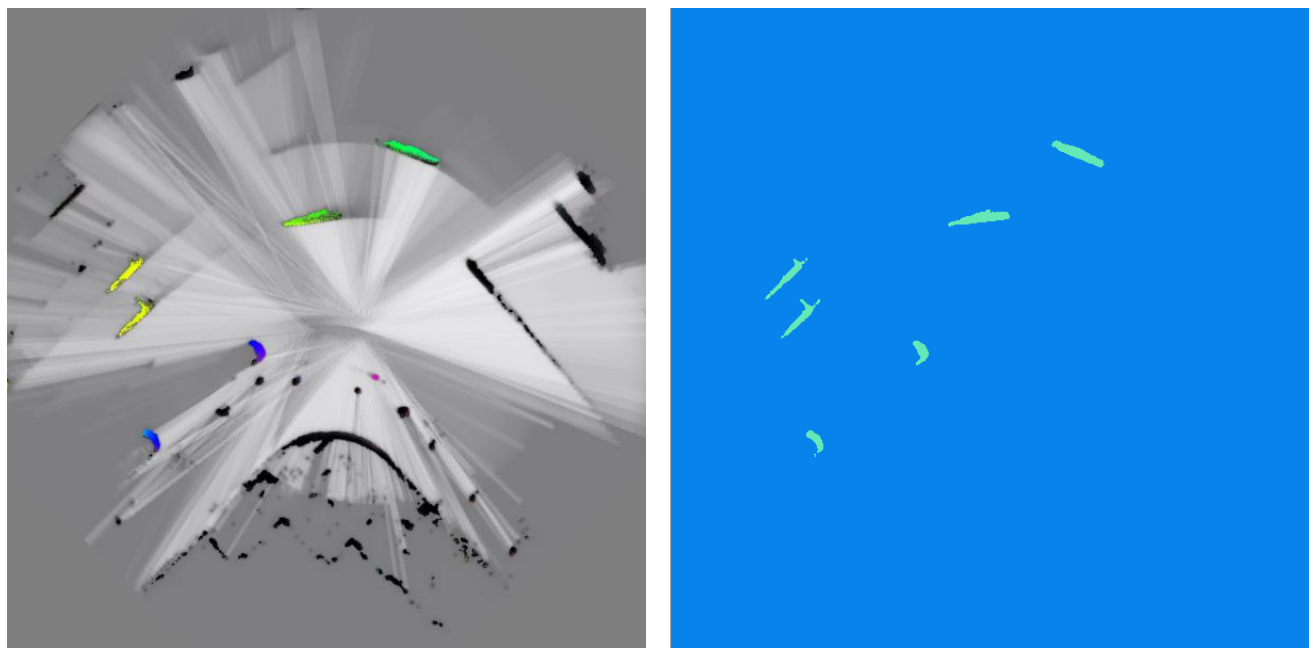

Fig. 1. Dynamic occupancy grid map (left) and dynamic objects detected with our approach (right).

## I. INTRODUCTION

The detection of obstacles in the surrounding of a vehicle is an elementary part of environment perception for self-driving cars and knowing whether an object is stationary or moving is particularly important to decide how to react to it. Stationary infrastructure for instance is important for localization, while a pedestrian approaching a cross-walk affects the planning component of the vehicle.

Occupancy grid maps are a well-established ([1], [2]) and popular environment model because of their simple structure and no assumptions on object models or shapes. Instead, they divide the environment into small grid cells and estimate cell properties, e.g. the occupancy state, based on sensor range data. Nuss et al. extended this to a dynamic occupancy grid map (DOG) that additionally estimates the velocity of individual grid cells by means of a particle filter [3]. Given such a method, the decision whether there is movement in a grid cell can in principle be done by comparing the estimated velocity against a threshold. However, in some cases DOG [3] falsely detects motion due to noisy measurements in cluttered areas or because of the aperture problem, which states that motion cannot be estimated correctly for one-dimensional spatial structures, such as curbs or crash barriers.

In this paper, we improve motion detection by taking into account this spatial structure of objects in the grid map, in addition to cell-level velocity estimates. We do so by training a Fully Convolutional Neural Network (FCN) that predicts whether individual grid cells are moving or not based on various cues within their surrounding. The rest of the paper is structured as follows: section II discusses related work, section III introduces the details of our approach, section IV outlines our semi-supervised generation of training data, section V shows results of our approach and how it compares to the reference method, and section VI describes the improvements of runtime performance of our approach.

[1]Mercedes-Benz R&D North America, Inc., Sunnyvale, USA
[2]Karlsruhe Institute of Technology, Karlsruhe, Germany
[3]FZI Research Center for Information Technology, Karlsruhe, Germany
Primary contact: florian.piewak@daimler.com

## II. RELATED WORK

Several publications exist about the detection of moving objects with various types of sensors. For example Vu et al. [4] implemented a dynamic obstacle detection by using a measurement grid of a laser sensor. Another approach from Asvadi et al. [5] uses a Velodyne sensor to generate a 2.5D grid map, where dynamic objects are detected by recognizing movements of grid cells over time. Franke et al. [6] estimate the velocity of individual point tracks to detect moving objects with a stereo camera and the Doppler effect of radar sensors can directly measure velocity, which is clearly very helpful for motion detection.

However, the aforementioned papers and principles are using only one specific sensor to distinguish dynamic and static objects. In contrast, Nuss et al. [3] proposed a dynamic occupancy grid map (DOG) that can combine information from multiple sensor types (e.g. lidar, radar, and camera) into one representation. In addition to occupancy and freespace, the DOG approach uses a particle filter to estimate the dynamic state of each grid cell in terms of absolute speed over ground. Given this DOG, Peukert et al. [7] cluster occupied grid cells to object proposals and mark them as dynamic or static based on the velocity estimates of all cells within the cluster. These clusters are then used to instantiate or update object-level tracks which incorporate knowledge such as vehicle dynamics [8] that are not taken into account in the DOG. In [7], only the particle velocities per grid cell are used to distinguish between dynamic and static clusters, which can cause problems with noise, in cluttered areas or due to the aperture problem.

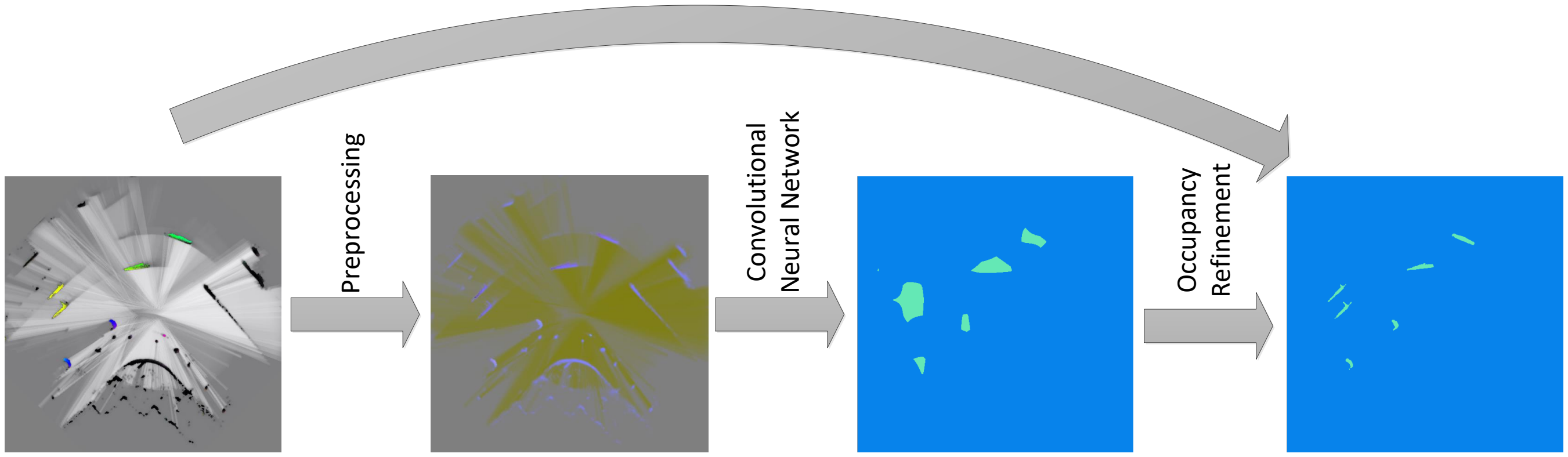

Fig. 2. Overview of our processing steps: the dynamic occupancy grid map (DOG) of [3] is pre-processed to extract a CNN-optimized input image. This input is then passed to a fully convolutional neural network (FCN) that marks dynamic cells in the grid map. The result inferred by the network is finally refined with the occupancy information of the DOG to remove false positives. The final output shows static labels in blue and dynamic labels in turquoise.

In the last few years Convolutional Neural Networks (CNNs) have rapidly improved and surpassed the performance of existing methods in various fields of research. The availibility of large amounts of data and rising computational power made training and execution of deeper networks practically feasible [9]. Approaches like the Hypercolumns of Hariharan et al. [10] use a CNN for feature generation. These features are produced over all the CNN layers by upscaling smaller layers and combining all the layers pixelwise. This produces a feature vector for each pixel, which can be used to train linear classifiers. A slightly different approach is the Region-based Convolutional Network from Girshick et al. [11] which creates classified bounding boxes instead of pixelwise predictions. There, the CNN is used to create feature vectors of generated region proposals which are then classified with a support vector machine (SVM). One of the problems of these approaches is that for each pixel or region proposal the classifier has to be executed. That produces a computational overhead. To avoid this problem, Long et al. [12] introduced Fully Convolutional Neural Networks (FCNs) in combination with deconvolutional layer, which were trained end-to-end for segmentation tasks. They transfered the standard AlexNet [13], VGG-net [14], and GoogLeNet [15] to FCNs and compared them based on the Pascal VOC 2012 dataset [16]. To create a fully convolutional network for pixelwise prediction, they used deconvolution layers to generate a pixel-wise predicion for each class. The deconvolution layers are implemented as up-sampling layers where the interpolation parameters are learned during training. This reduces the computational overhead so that for a pixelwise prediction the net has to be executed only once. To refine the segmentation result, several deconvolution layers combined with convolutional layers were stacked. This results in more detailed segmentation structures at the output layer. Based on this approach, several extensions and applications were developed, e.g. optical flow calculation by Fischer et al. [17] or a refinement with Conditional Random Fields (CRFs) [18], [19] to produce more detailed structure of the segmented objects. In this work, we use the standard FCN of Long et al. [12].

## III. APPROACH

The goal of our work is to create a pixelwise classification of a dynamic occupancy grid map (DOG), to support detection of moving obstacles. We use the DOG of [3] as input, but our approach can also be applied to other grid maps. Our target set consists of two classes, dynamic (moving objects) and static (non-moving objects), where background is part of the static class. The classification process is divided into three steps, as shown in Figure 2. First, the DOG is preprocessed to obtain an image that is easier to analyze for the CNN. Based on this image the CNN infers the pixelwise classification. Finally, the intersection between the segmented image and an image of occupied grid cells is calculated, to produce sharper labeling borders and rejects false positives which appear by labeling non-occupied grid cells as dynamic.

To achieve the classification task based on a grid map image, Fully Convolutional Neural Networks (FCNs) are used to reduce the computationally expensive task of sliding window approaches. Based on Long et al. [12], we convert two popular network structures (VGG-net [14] and Alexnet [13]) to FCNs by adding deconvolutional layers (also called upscaling layers). This produces a two-dimensional output of the network and creates the desired pixelwise classification, where input and output image size of the network are identical. For initialization of the CNN weights, pre-trained networks are used to reduce the training time. These pre-trained networks were trained on the public Imagenet [20] dataset, which contains colored images of different categories. Although this input is quite different to our grid map data, this procedure helps to reduce training time and improve performance, because the filters of the lower convolutional layers contain rather generic shape representations that also appear in grid maps, i.e. blobs, lines, and corners. However, the transfer of the pre-trained networks to a new scope still requires several adaptations and parameter optimizations. In the following, we discuss the three most important aspects we analyzed: (1) how to best prepare the grid map input data for the CNN, i.e. the preprocessing step, (2) which network structure is best suited for the problem of cell-level motion detection, and (3) how to balance the strong class bias between background (static) and foreground (dynamic) cells. For further optimization aspects of our work, the reader is referred to [21].

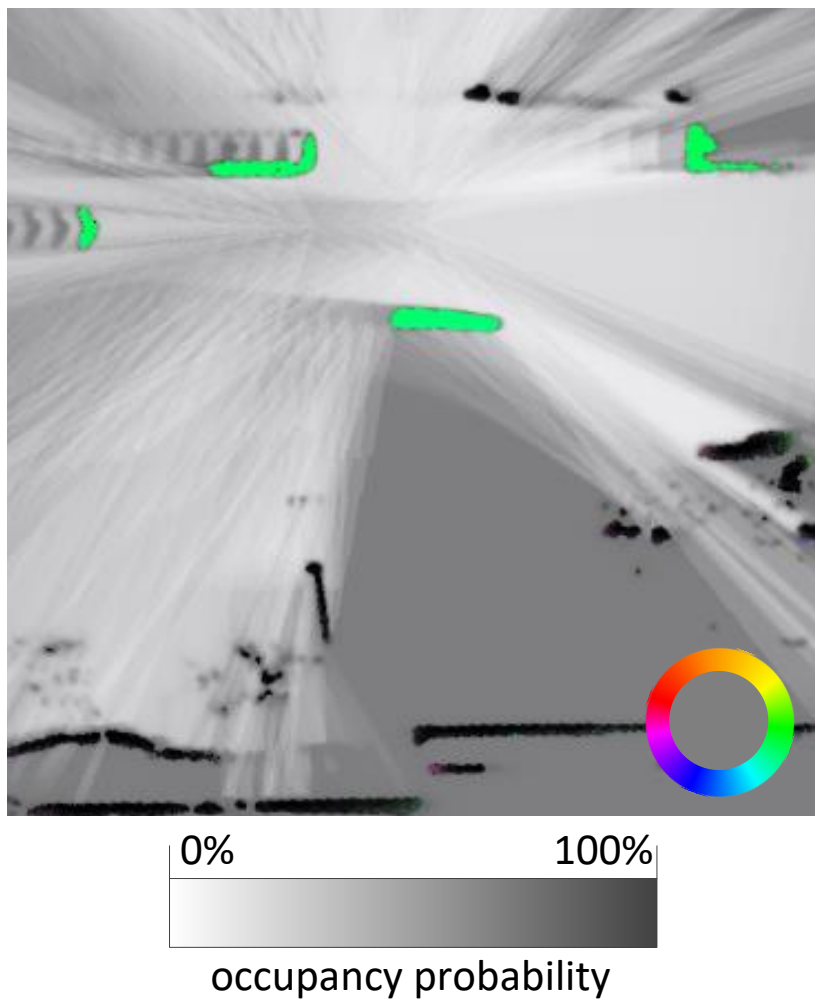

Fig. 3. Example of a dynamic occupancy grid map (DOG): the orientation of the velocity is encoded in a color (see the circle) and the occupancy probability is encoded with gray scale values.

### A. Input configuration

The DOG [3] estimates occupancy and freespace, as well as a velocity distribution for each grid cell by means of a particle filter, where the occupancy value is proportional to the weight of all particles in a grid cell. Occupancy and freespace are encoded with a scalar value between 0.0 and 1.0, where 0.0 represents freespace and 1.0 represents an obstacle. If no freespace information can be inferred, the value is only between 0.5 and 1.0, as shown in Figure 3. Each particle has a dynamic state in the form of velocity in $x$ and $y$ direction. For one grid cell, the mean velocity and the variance of the velocities of all particles can be calculated, which represents the uncertainty of the velocity estimate. Based on that, a Mahalanobis distance $m$ of the mean overall velocity $\vec{v}_{overall}$ to the static reference $\vec{v} = \vec{0}$ can be calculated, as mentioned in [3]:

$$m^2 = \vec{v}_{overall}^T \Sigma^{-1} \vec{v}_{overall} \ , \tag{1}$$

where $\Sigma$ represents the covariance matrix of the velocities.

The structure of the used CNN is designed based on public network structures such as VGG-net [14] and Alexnet [13]. To be able to use weights of a pre-trained model, we encode the aforementioned grid map parameters into three channels to obtain a regular RGB image. The following combinations of input encodings have been evaluated (B/G/R):

1) $Occ_{free}$ / $v_x$ / $v_y$
2) $Occ_{free}$ / $v_{x,norm}$ / $v_{y,norm}$
3) $Occ_{nonfree}$ / $v_{x,norm}$ / $v_{y,norm}$
4) $Occ_{free}$ / $v_{overall}$ / $Var_{overall}$
5) $Occ_{free}$ / $v_{overall}$ / $m$

In every combination the occupancy probability is included. The reason for this is that the CNN should use this occupancy information to include shape recognition of the obstacles for the prediction. The third combination excludes the freespace information to analyze the influence of freespace information. All the other parameters are used to represent the movement of grid cells. Besides the velocity itself, the normalized velocities $v_{x,norm}$ and $v_{y,norm}$ can be provided. The normalized velocity in $x$ and in $y$ direction can be calculated accordingly, where the normalized velocity in $x$ direction is defined as

$$v_{x,norm} = \frac{v_x}{\sqrt{Var(v_x)}} \ . \tag{2}$$

The normalization with the variance should help to distinguish between clutter and real moving objects. This is caused by the DOG: grid cells, which belong to clutter, contain particles with different velocities. Additionally, in combination four and five, the parameters for the uncertainty are provided in a separate input channel. One of these parameters is the overall variance, which is calculated with

$$Var_{overall} = Var(v_x) + 2Cov(v_x, v_y) + Var(v_y) \ . \tag{3}$$

Another parameter is the mahalanobis distance (see Equation (1)). The problem of these combinations is the restriction of three input channels. This causes the velocities to be combined to the overall velocity $\vec{v}_{overall}$.

### B. Neural network structure

Based on Long et al. [12], we extend two public network stuctures in a way that the inferred results are upscaled to the original input resolution. This is done by transferring weights of fully connected layers to convolutional layers and introducing an additional deconvolutional layer. Caused by the network structure, the single deconvolutional layer has a stride of 32 to upscale the image to the input size. This limits the resolution of details in the segmentation because the inner representation is 32 times smaller than the final output. To improve this, Long et al. [12] introduced a stepwise deconvolution, called deep jet. There, each deconvolution step is fused with lower layers to recombine higher semantic information with lower shape information. The last deconvolution layer then uses a smaller stride, which produces a finer segmentation output. The two network structures used in this paper are VGG-net [14] and Alexnet [13]. VGG-net is deeper than Alexnet, i.e. it contains more layers. This results in longer execution time, but also in higher accuracy, c.f. [12]. For both networks, we evaluate three different variants, where we use a single deconvolutional layer but vary the stride parameter of this layer. This is indicated later as FCN-Xs (for VGG-net) and Alex-Xs (for Alexnet), respectively, where X corresponds to the applied stride value.

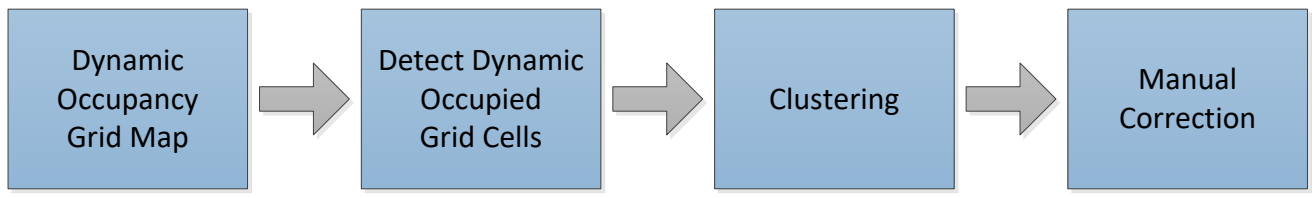

Fig. 4. Overview of semi-automatic labeling toolchain.

### C. Weight matrix to handle class bias

As shown in Figure 1 and Figure 3, the number of static pixels (including the background) is much higher than the number of dynamic pixels. That can cause problems with the backpropagation algorithm, which puts much more emphasis on the static class. To compensate this effect, a weight matrix $C$ is introduced in the cost function $J(\theta)$ of the multinomial logistic loss (for further details the reader is referred to [22]):

$$J(\theta) = -\left[\sum_{i=1}^{m}\sum_{k=1}^{K} C^{\left(y^{(i)}\right)} 1\{y^{(i)} = k\} \log Q\left(i, k, \theta\right)\right] , \tag{4}$$

where $Q(i, k, \theta)$ is defined as

$$Q(i, k, \theta) = P\left(y^{(i)} = k | x^{(i)}; \theta\right) \tag{5}$$

and $C$ contains a weight for each labeled class:

$$C = \left[c^{(1)} c^{(2)} \dots c^{(K)}\right] . \tag{6}$$

## IV. TRAINING DATA

We created data to train and evaluate our approach by recording a sequence in Sunnyvale, CA, that stretches over 4.3 miles and contains various different traffic situations. The recording includes six-lane streets, with a speed limit of 45 miles per hour, passing maneuvers of the recording vehicle and from other vehicles, stopping at traffic lights, where several vehicles crossed the street, a limited traffic zone as well as a parking lot of a shopping center, where pedestrians, bicycles, and vehicles with a lower speed were recorded. This results in a certain diversity of the data to facilitate generalization of urban areas.

To annotate the data we follow a semi-automated procedure, where we use DBSCAN [23] to automatically generate motion clusters based on the existing velocity estimates of the DOG, following [7]. These clusters are then converted to polygons describing the convex hull of each cluster. In the last step, false positive and false negative clusters are corrected manually with a polygon-based annotation tool. The annotation procedure is depicted in Figure 4.

During this process, not all of the data is used. Especially if the recording vehicle had to wait on a traffic light for a left turn or if the recorded images are too similar, the annotation is skipped. Overall, 3 457 images were labeled with this semi-automatic approach. Given that the ego-vehicle is always located in the center of our grid map, we augmented the data by rotating the grid map in 10-degree steps to yield rotation invariance of the algorithm and more data for training.

The data is split in different datasets to prevent overfitting and evaluate the generalization. A split of roughly $80\% - 10\% - 10\%$ for training, validation, and test data is chosen, where the training set is used for training the CNN, the validation set is used to optimize the parameters (see Section III) and the test set is used to evaluate the CNN against the baseline method of Nuss et al. [3].

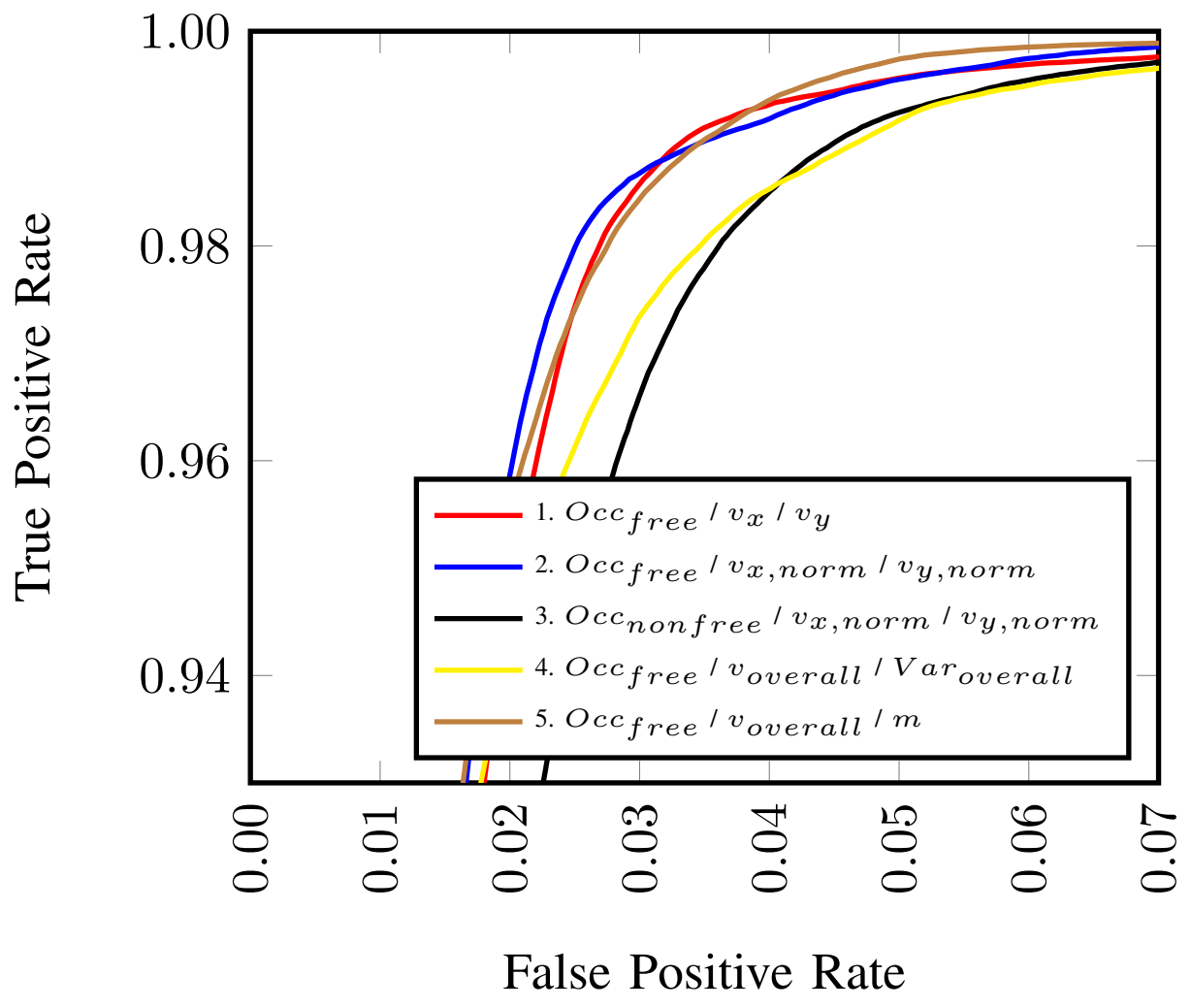

Fig. 5. ROC curves of different input configurations.

## V. EXPERIMENTS

The results of the experiments are evaluated with ROC curves. As mentioned in Section III-C, the number of background grid cells in the DOG is much higher than the number of occupied grid cells. As we focus on differentiating moving from stationary objects, we only consider occupied grid cells (with an experimentally determined occupancy value greater than 0.6) in our ROC curve evaluation. Doing so has the same effect as our proposed occupancy refinement and provides a more realistic performance estimate.

The results in Section V-A, V-B, and V-C are generated on the validation set, while the final comparison with our baseline in Section V-D is performed on the test set.

### A. Input configuration

In this subsection, the results of the input combinations of Section III-A are discussed. Figure 5 shows the ROC curves of the input configurations. In general, the combinations with separation of velocities in $x$ and $y$ direction (comb. 1 and 2) and the combination with the mahalanobis distance (comb. 5) reach a better result than the combination with the separate overall variance (comb. 4) and the combination without the freespace information (comb. 3). That proves on the one side that the split of the normalized velocity of different direction into the overall velocity and the overall variance creates no information gain for the CNN. On the other side, it proves that freespace information is quite important for recognizing moving obstacles. This can be explained with an example in Figure 6. There it is easy to distinguish between a corner of a wall, where at the inside of the L-shape is freespace information, and a possible vehicle. Without the freespace information, the two objects have the same appearance.

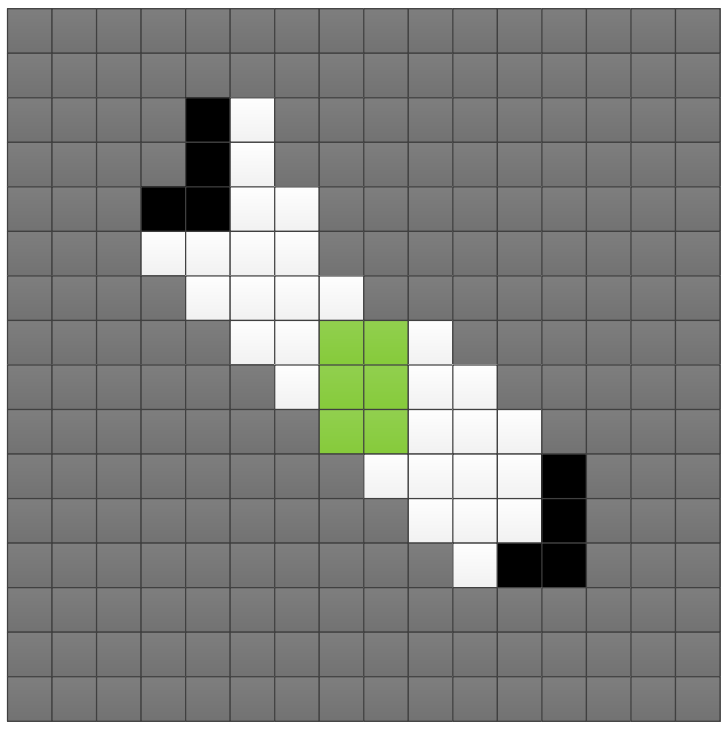

Fig. 6. Example of the recognition of a vehicle (upper left L-shape) and a wall (lower right L-shape) within a DOG by using the freespace information: the green object in the middle is the recording vehicle.

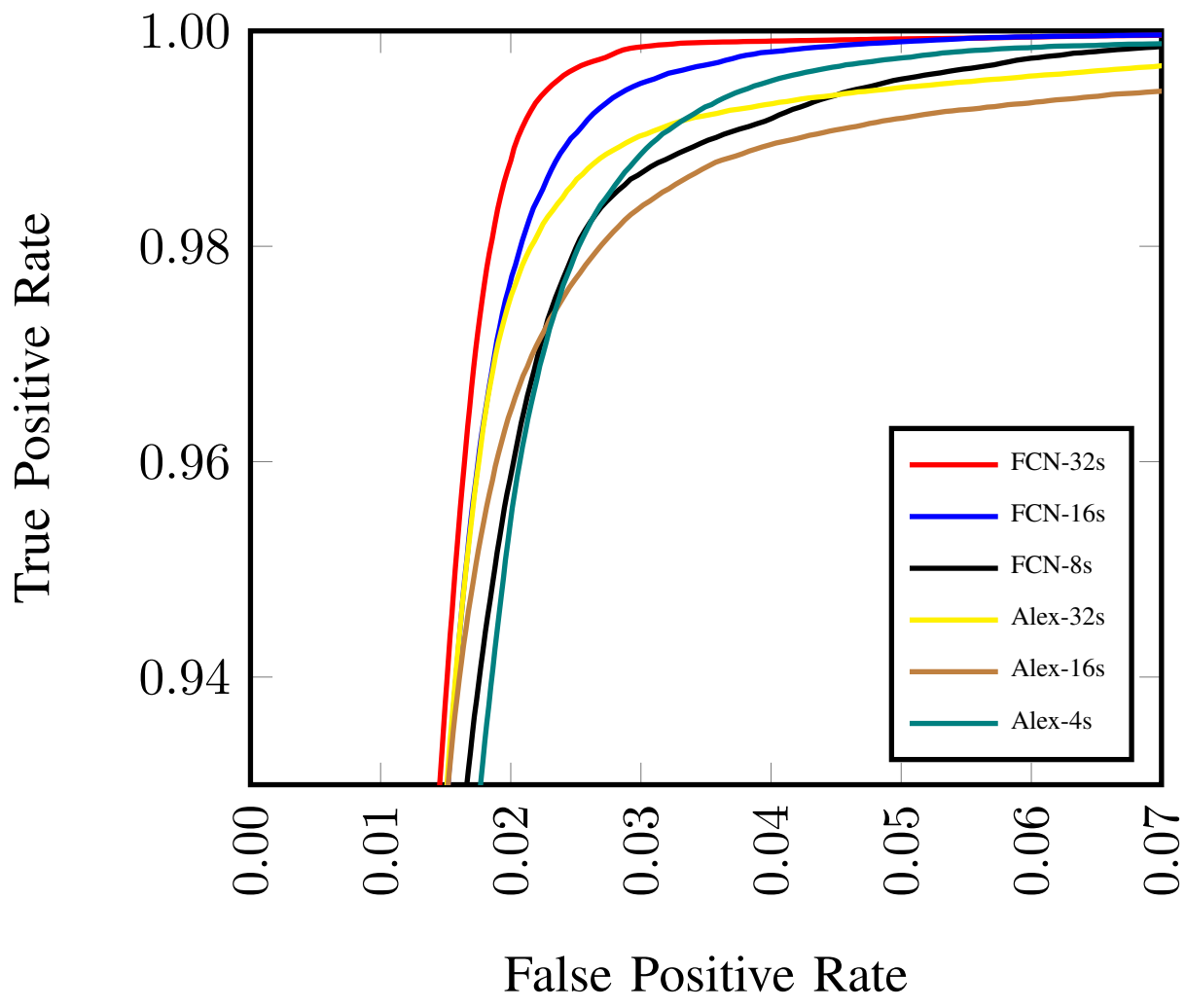

Fig. 7. ROC curves of the VGG-net with different deep jet configuration.

Some smaller differences can be seen between the combination with the mahalanobis distance (comb. 5) and the normalized and unnormalized approaches (comb. 1 and 2). There, the mahalanobis distance has a slightly lower performance than the other two approaches. A suggestion for the reason is the fusion of the velocity in $x$ and in $y$ direction to the overall velocity. There the information of the moving direction is lost. So some specific motion model cannot be calculated.

The combination with the normalized velocities in $x$ and $y$ direction (comb. 2) creates the best result. It is slightly better than the combination with the unnormalized velocities. That can be explained with the appearance of clutter. It can be observed, that clutter can reach a high velocity but has also a high variance. If the velocities are normalized, it is easier to mark motion clutter as static background.

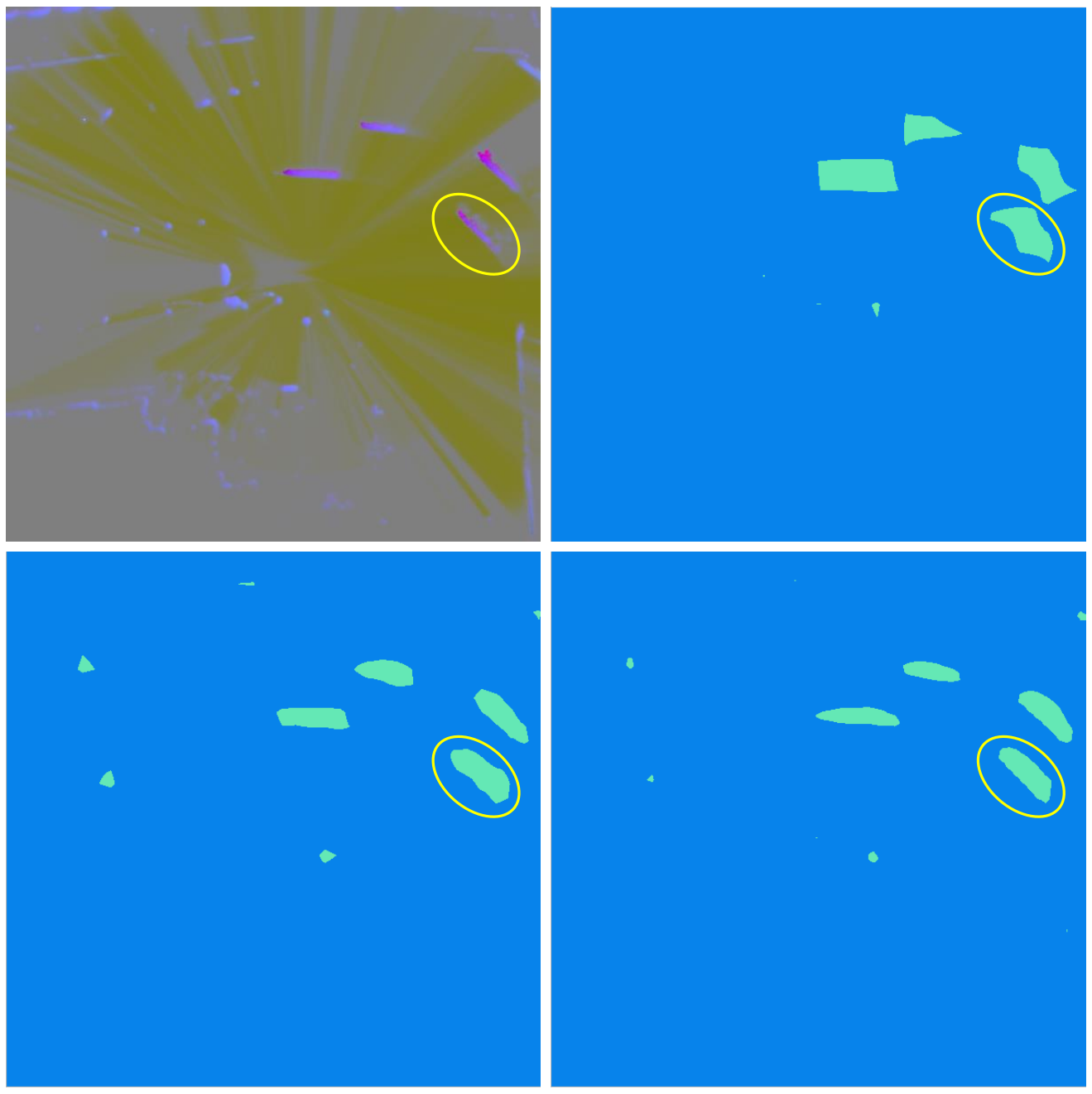

Fig. 8. Example of the result of deep jet of VGG-net: top left: input image, top right: output of the FCN-32s, bottom left: output of the FCN-16s, bottom right: output of the FCN-8s. In general, the smaller stride at the last layers (for example at FCN-8s or FCN-16s) produces finer details of objects, but increases also the number of false positive clusters.

### *B. Neural network structure*

In this subsection, the structure configuration of Section III-B is evaluated. The results are shown in Figure 7. There, the FCN-32s is the best network configuration. This is an interesting result, because Long et al. [12] showed the opposite in their setting. They describe that the FCN-8s version creates finer details within the segmentation, what leads to a better result. In fact the FCN-8s variant also procudes more details in our setting, as shown in Figure 8. There, the segmentation output of the FCN-32s, FCN-16s, FCN-8s, and the input data is provided. The FCN-32s created much coarser results than the FCN-8s, but it also creates less false positive clusters. Additionally, the filter sizes of the last deconvolutional layer can be recognized at the marked vehicle (yellow ellipse). There, the up-sampling artifacts can be recognized, which become smaller with a smaller stride as with the FCN-16s or the FCN-8s. The output of the CNN is not used directly for segmentation. As mentioned in Section III, the output of the CNN is refined with the occupancy information of the DOG by setting the label to static when the occupancy value falls under 0.6. With this refinement included, the FCN-32s generates a better result than the FCN-16s or FCN-8s due to less false positives. That means our refinement with the occupancy information of the DOG is better for grid maps than the refinement with deep jet.

As mentioned in Section III-B, the resulting quality should also be put into perspective with execution time. For this purpose, we also evaluate the smaller Alexnet as comparison.

| Network Structure | Preprocessing and Postprocessing | Network Inference Time |
|---|---|---|
| FCN-32s | $\approx 100ms$ | $\approx 160ms$ |
| Alex-32s | | $\approx 20ms$ |

TABLE I

EXECUTION TIME OF DIFFERENT NETWORK STRUCTURES.

The results are shown as ROC curves in Figure 7 as well. The ROC curves show that the Alexnet generally has a lower performance than the FCN-Xs networks. This is as expected due to the smaller structure of the network.

The execution time of the Alexnet is much smaller, what can be seen in Table I. There, the execution time is divided into two parts. The first part is the pre-processing and post-processing time, which is needed to create the specific input image for the CNN (see Section III-A) and to refine the output with the occupancy information of the DOG. The second part of the execution time is the processing of the CNN itself, which is already highly optimized on the GPU. The shorter execution time has to be seen in contrast to the quality of the output. All performance evaluations were executed on an NVIDIA GeForce GTX Titan X (Maxwell) in combination with an Intel Core I7 ($6^{th}$ generation) by using Caffe as CNN framework [24].

### *C. Weight matrix*

In this subsection, the weight matrix configuration of Section III-C is evaluated. To define the weight matrices for evaluation, the ratio between moving grid cells and non-moving grid cells of the validation dataset was calculated, which is nearly $1:200$. For this reason the following factors $c^{(1)}$ for the moving grid cells were defined:

$$c^{(1)} \in \{1, 20, 40, 60, 80, 100, 120, 140, 160, 180, 200\} \ . \quad (7)$$

Note that the factor for the non-moving grid cells is set to

$$c^{(2)} = 1 \ . \quad (8)$$

As a result of the weight matrix adaption, the factor $c^{(1)} = 1$ has the lowest performance and that the quality of the result increase with a growing factor. After a factor of $c^{(1)} = 40$ the quality decreases. That result demonstrates the improvement of the quality by using a factor to solve the imbalance of the labeled data. Especially when the imbalance is as high as in our dataset, it is important to weight the labels. Otherwise the CNN would focus on the background.

### *D. Comparison to baseline*

For the final comparison to the baseline method we employ the best network configuration we found during the previous experiments. The CNN is evaluated against the baseline method introduced by Nuss et al. [3], which distinguishes dynamic and static grid cells by thresholding the Mahalanobis distance $m$ between the estimated two-dimensional velocity distribution $p(\vec{v})$ in a cell and the velocity $\vec{v} = \vec{0}$.

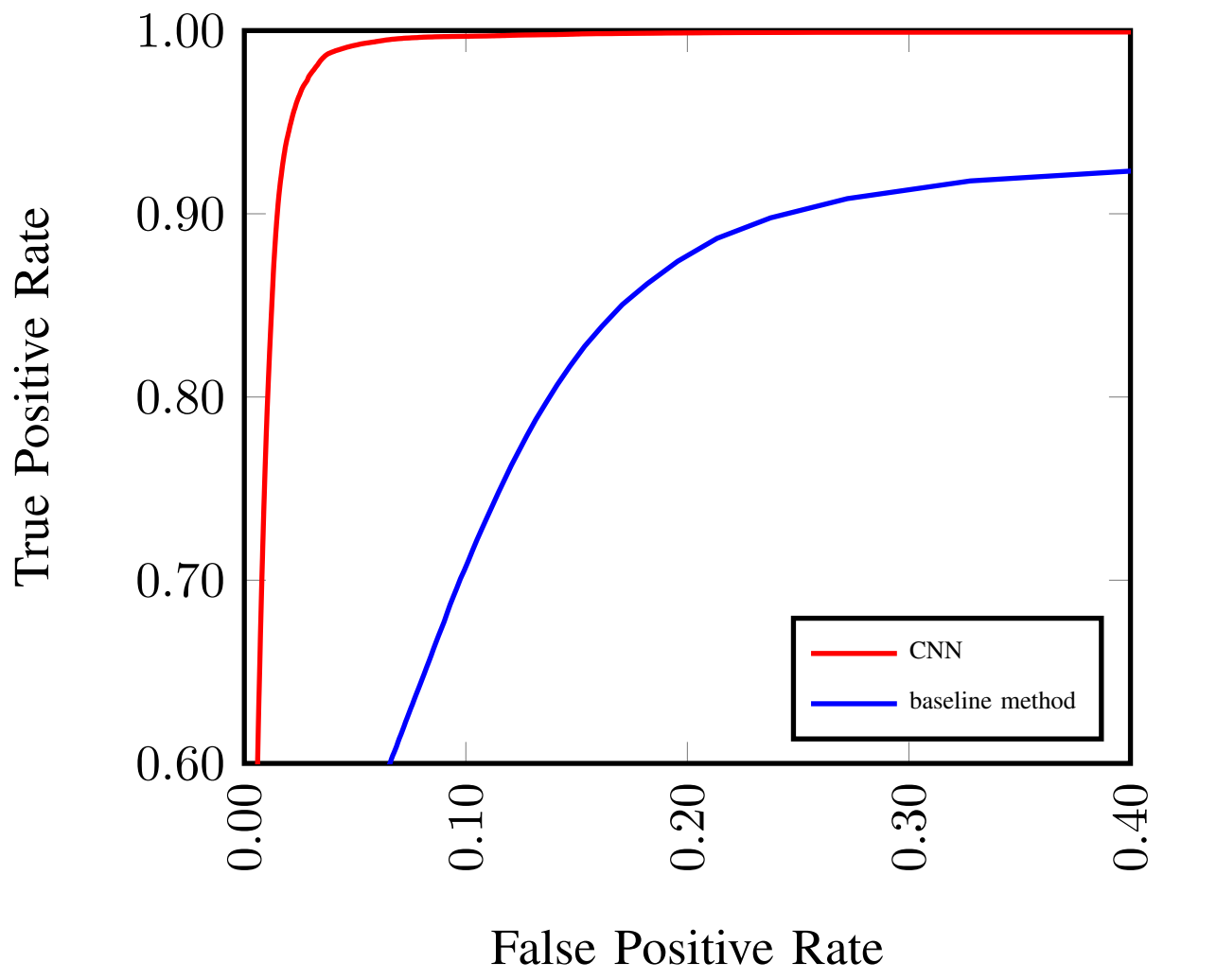

Fig. 9. ROC curves of the final comparison with Nuss et al. [3] as baseline method.

The results of this comparison are shown in Figure 9. In terms of equal error rate (EER), the baseline delivers $83.9\%$. In contrast, our proposed methods yields $97.2\%$, resulting in a relative improvement of $15.9\%$, which is a clear indicator that structural information in the occupancy grid map helps to distinguish actual moving objects from stationary infrastructure and motion clutter.

The baseline method creates a decision for each grid cell independently. That results in a complete rotation invariance of objects. The CNN internally uses filters at the convolutional layer of a specific size, which results in a specific field of view. That can affect the rotation invariance of a CNN. For this reason the input images are rotated during training as described in Section IV. To examine the rotation invariance in this section, the test input images are rotated as well, where one test sample is shown in Figure 10. The figure shows that the precision, recall and the accuracy do not really change over the different rotation angles. This shows that the CNN is also rotation invariant. Additionally it can be recognized, that the precision and the recall have the same curve progression in a distance of 90 degrees. That is caused by the filters of the convolutional layers, which are squared. By rotating the input image, the squared filter changes the field of view caused by the corners of the square. To reduce this and enforce the rotation invariance, a circular filter shape could be used in future work.

## VI. RUNTIME

Based on the Alex-32s from Section V, we focused on improving the runtime performance of our approach. In order to cut down execution time, we reduced the input resolution from 545x545 to 273x273, moved all pre-processing to the GPU, reduced the filter sizes in the convolutional layers of the network and removed one convolution layer plus pooling stage. These changes result in an improvement of the execution time to around 2 ms for the pre- and post-processing and around 8 ms for the network inference time. At the same time, we can still keep a relative improvement of 14.5% compared to the baseline method.

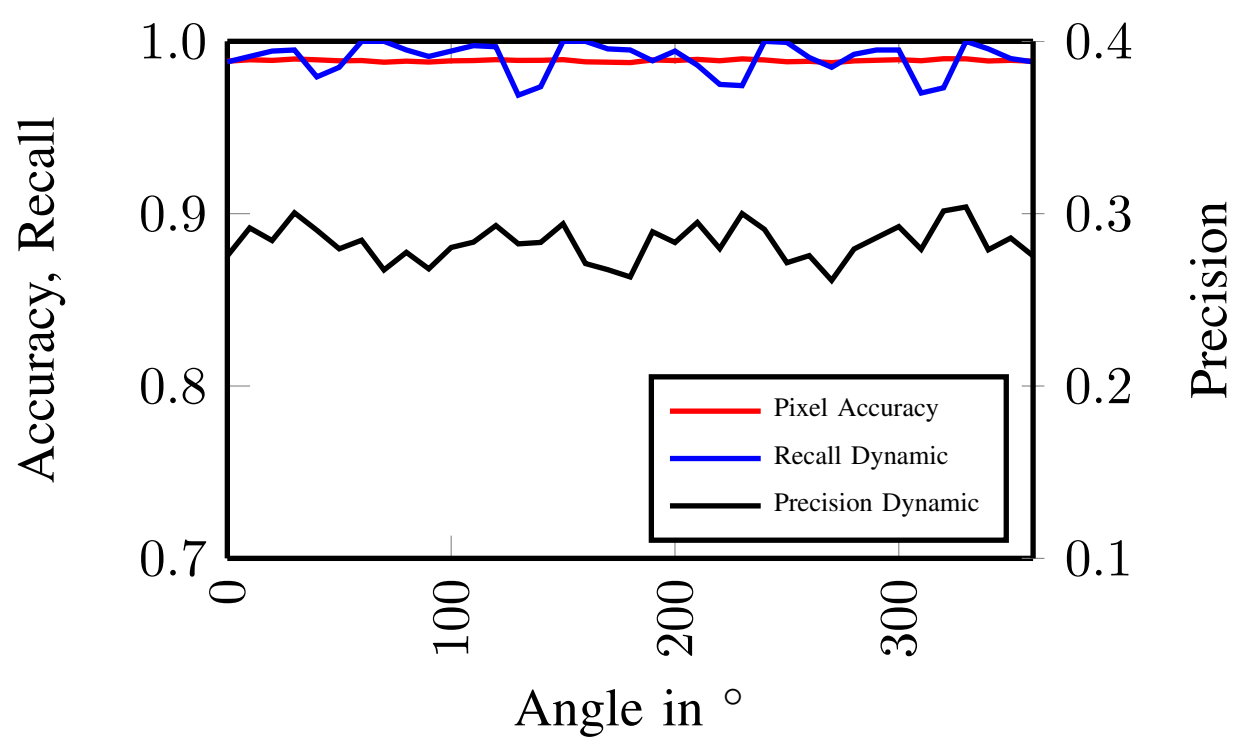

Fig. 10. Evaluation of orientation invariance by rotating one input image of the test dataset.

## VII. CONCLUSION

In this paper we introduced a novel approach to detect moving obstacles in grid maps based on a deep convolutional neural network. Compared to existing approaches, our network takes into account the spatial occupancy and freespace structure in the grid map and thus can better differentiate real moving objects from motion clutter that arises from noisy measurements in the grid map or due to the aperture problem. Compared to the baseline approach of [3], our methods yields a 15.9% increase. Our real-time variant still yields a 14.5% increase with just 10 ms execution time. The outcome of our approach can be used e.g. to create robust candidates for high-level object tracker.

Further extensions of our work can be achieved by adapting the network architecture to a different number of input channels, which should increase the performance by using more than three channels. Additionally the classification task can be extended to a multi-class problem by detecting the type of moving objects (e.g. vehicle, pedestrian). This can further improve high-level object tracker by adapting the motion model to the detected object type.

## ACKNOWLEDGMENT

The authors would like to thank Lisa Schermuly for the tedious manual annotation effort that went into creating the dataset. This research was supported by Mercedes-Benz Research & Development North America (MBRDNA), Inc. and the FZI Research Center for Information Technology, Germany.